\documentclass[11pt,a4paper]{article}
\usepackage[hyperref]{emnlp-ijcnlp-2019}
\usepackage{times}
\usepackage{latexsym}
\usepackage{graphicx}
\usepackage{arydshln}
\usepackage{multirow}
\usepackage{amsfonts}
\usepackage{subfig}
\usepackage{CJK}

\usepackage{url}

\aclfinalcopy 

\usepackage{color}
\definecolor{zptu}{RGB}{18, 141, 21}

\title{Multi-Granularity Self-Attention for Neural Machine Translation}

\author{Jie Hao\thanks{~~Work done when interning at Tencent AI Lab.}\\\normalsize Florida State University\\{\normalsize \tt haoj8711@gmail.com} \And
Xing Wang\\\normalsize Tencent AI Lab\\{\normalsize \tt  brightxwang@tencent.com} \AND
Shuming Shi\\\normalsize Tencent AI Lab\\{\normalsize \tt shumingshi@tencent.com} \And 
Jinfeng Zhang\\\normalsize Florida State University\\{\normalsize \tt jinfeng@stat.fsu.edu} \And
Zhaopeng Tu\\\normalsize Tencent AI Lab\\{\normalsize \tt zptu@tencent.com}
}

\date{}

\begin{document}
\begin{CJK}{UTF8}{gkai}
\maketitle
\begin{abstract}

Current state-of-the-art neural machine translation (NMT) uses a deep multi-head self-attention network with no explicit phrase information. However, prior work on statistical machine translation has shown that extending the basic translation unit from words to phrases has produced substantial improvements, suggesting the possibility of improving NMT performance from explicit modeling of phrases.
In this work, we present {\em multi-granularity self-attention} (\textsc{Mg-Sa}): a neural network that combines multi-head self-attention and phrase modeling. 
Specifically, we train several attention heads to attend to phrases in either n-gram or syntactic formalism.
Moreover, we exploit interactions among phrases to enhance the strength of structure modeling -- a commonly-cited weakness of self-attention.
Experimental results on WMT14 English-to-German and NIST Chinese-to-English translation tasks show the proposed approach consistently improves performance. 
Targeted linguistic analysis reveals that \textsc{Mg-Sa} indeed captures useful phrase information at various levels of granularities.

\end{abstract}

\section{Introduction}

Recently, \textsc{Transformer}~\cite{Vaswani:2017:NIPS}, implemented as deep multi-head self-attention networks (\textsc{San}s), has become the state-of-the-art neural machine translation (NMT) model in recent years.
The popularity of \textsc{San}s lies in its high parallelization in computation, and flexibility in modeling dependencies regardless of distance by explicitly attending to all the signals.


More recently, an in-depth study~\cite{raganato2018analysis} reveals that \textsc{San}s generally focus on disperse words and ignore continuous phrase patterns, which have proven essential in both statistical machine translation~\citep[{SMT,}][]{koehn2003statistical,chiang:2005:acl,Liu:2006:ACL} and NMT~\cite{eriguchi:2016:acl,wang:2017:emnlp,yang:2018:emnlp,czhaophrase}.

To alleviate this problem, in this work we propose {\em multi-granularity self-attention} (\textsc{Mg-Sa}), which offers \textsc{San}s the ability to model phrases and meanwhile maintain their simplicity and flexibility.
The starting point for our approach is an observation: the power of multiple heads in \textsc{San}s is not fully exploited. For example,~\newcite{Li:2018:EMNLP} show that different attention heads generally attend to the same positions, and ~\newcite{Voita:2019:ACL} reveal that only specialized attention heads do the heavy lifting while the rest can be pruned. Accordingly, we spare several attention heads for modeling phrase patterns for \textsc{San}s.

Specifically, we use two representative types of phrases that are widely-used in SMT models: {\em n-gram phrases}~\cite{koehn2003statistical} to use surface of adjacent words, and {\em syntactic phrases}~\cite{Liu:2006:ACL} induced from syntactic trees to represent well-formed structural information. We first partition the input sentence into phrase fragments at different levels of granularity. For example, we can split a sentence into 2-grams or 3-grams. Then, we assign an attention head to attend over phrase fragments at each granularity. In this way, \textsc{Mg-San}s provide a lightweight strategy to explicitly model phrase structures.
Furthermore, we also model the interactions among phrases to enhance structure modeling, which is one commonly-cited weakness of \textsc{San}s~\cite{Tran:2018:EMNLP,Hao:2019:NAACL}.

We evaluate the proposed model on two widely-used translation tasks: WMT14 English-to-German and NIST Chinese-to-English. Experimental results demonstrate that our approach consistently improves translation performance over strong \textsc{Transformer} baseline model~\cite{Vaswani:2017:NIPS} across language pairs, while speeds marginally decrease. 
Analysis on multi-granularity label prediction tasks reveals that \textsc{Mg-Sa} indeed captures and stores the information of different granularity phrases as expected.

\section{Background}
\paragraph{Multi-Head Self-attention}
Instead of performing a single attention, Multi-Head Self-attention Networks (\textsc{Mh-Sa}), which are the defaults setting in \textsc{Transformer}~\cite{Vaswani:2017:NIPS}, project the queries, keys and values into multiple subspaces and performs attention on the projected queries, keys and values in each subspace. In the standard \textsc{Mh-Sa}, it jointly attends to information from different representation subspaces at different positions. Specifically, \textsc{Mh-Sa} transform input layer ${\bf H} = {h_{1}, ..., h_{n}} \in \mathbb{R}^{n \times d}$ into $h$-th subspace with different linear projections:
\begin{equation}
  {\bf Q}^h, {\bf K}^h, {\bf V}^h   = {\bf H}{\bf W}^h_{Q}, {\bf H}{\bf W}^h_{K}, {\bf H}{\bf W}^h_{V},
\end{equation}
where $\{{\bf Q}^h, {\bf K}^h, {\bf V}^h\} \in \mathbb{R}^{n \times d_{h}}$ are respectively the query, key, and value representations of the $h$-th head, $\{{\bf W}^h_{Q}, {\bf W}^h_{K}, {\bf W}^h_{V} \in \mathbb{R}^{d\times d_{h}} \}$
denote parameter matrices associated with the $h$-th head, $d$ and $d_{h}$ represent the dimensionality of the model and $h$-th head subspace. Moreover, $N$ attention functions are applied to generate the output states $\{{\bf O}^{1}, ..., {\bf O}^{N}\}$ in parallel, among them: 
\begin{equation}
    {\bf O}^{h} = \textsc{Att}({{\bf Q}^{h}}, {\bf K}^{h}) \ {\bf V}^{h}. 
\end{equation}
Finally, the output states are concatenated to produce the final state. Here \textsc{Att} denotes attention models, which can be implemented as either additive attention or dot-product attention. In this work, we use dot-product attention which is efficient and effective compared with its additive counterpart~\cite{Vaswani:2017:NIPS}: 
\begin{eqnarray}
    \textsc{Att}({{\bf Q}^{h}}, {\bf K}^{h})= softmax(\frac{{\bf Q}^{h}{\bf K}^{h^{T}}}{\sqrt{d_{h}}}),
\end{eqnarray}
where $\sqrt{d_{h}}$ is the scaling factor.

\paragraph{Motivation}
We demonstrate our motivation from two aspects. On the one hand, the conventional \textsc{Mh-Sa} model the individual word dependencies, in such scenario the query directly attends all words in memory without considering the latent structure of the input sentence. We argue that self-attention can be further improved by taking phrase pattern into account.  
On the other hand, recent study~\cite{Vaswani:2017:NIPS} implicitly hint that attention heads are underutilized as increasing number of heads from 4 to 8 or even 16 can hardly improve the translation performance. Several attention heads can be further exploited under specific guidance to improve the performance~\cite{Strubell:2018:EMNLP}. We expect the inductive bias for multi-granularity phrase can further improve the performance of \textsc{San}s and meanwhile maintain its simplicity and flexibility.

\section{Multi-Granularity Self-Attention}
We first introduce the framework of the proposed \textsc{Mg-Sa}. Then we describe the approaches of generating multi-granularity representation on a certain granularity representation. Finally, we introduce the training objective of our model with auxiliary supervision.  

\subsection{Framework}
\begin{figure*}[t]
    \centering
    \subfloat[Syntactic phrase partition]{
    \includegraphics[width=0.9\textwidth]{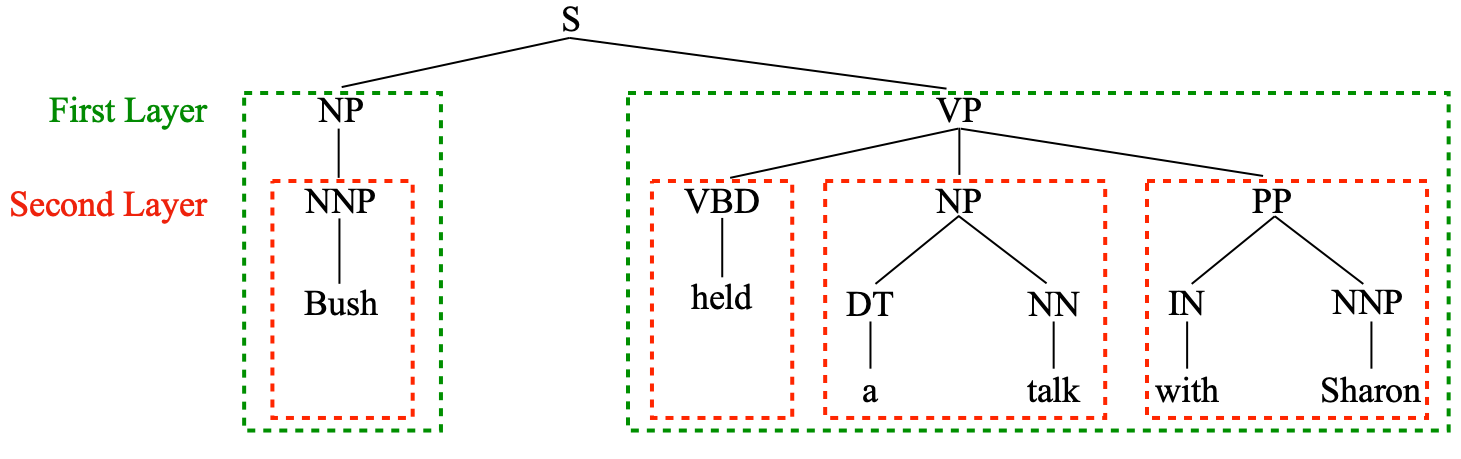}
    } \\
    \subfloat[{\em Multi-Granularity Self-Attention} on syntactic phrase partition]{
    \includegraphics[width=0.9\textwidth]{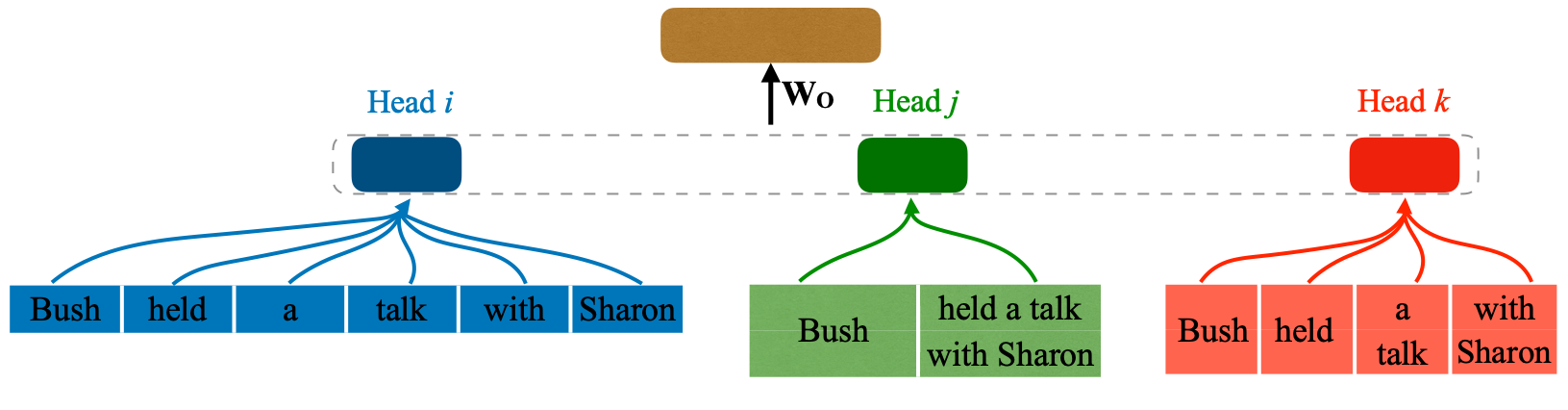}
    }
    \caption{Illustration of the proposed \textsc{Mg-Sa} model for syntactic phrase partition. In this example, we partition the sentence with top two layers in the constituent parse tree and obtain the syntactic phrase partitions (``Bush'', ``held a talk with Sharon''), (``Bush'', ``held'', ``a talk'', ``with Sharon''). Under the syntactic partition, multi-head attention in \textsc{Mg-Sa} attends the phrase memory (heads $\mathit j$ and $\mathit k$) as well as the conventional word memory (head $\mathit i$). The approach of phrase memory representation is described in Section 3.2. Best viewed in colour.} 
    \label{fig:mg-san}
\end{figure*}

The proposed \textsc{Mg-Sa} aims at improving the capability of \textsc{Mh-Sa} by modeling both word and phrase. We introduce various phrase granularity over the conventional word-level memory to generate phrase level memory.

Specifically, we first transform the input layer $\bf H$ to a phrase level memory by function $F_{h}$ in certain attention head:
\begin{eqnarray}
    {\bf H}_{g} &=& F_{h}(\bf H), \label{eq:fh}
\end{eqnarray}
where ${\bf H}_{g}$ is the generated phrase level memory, $h$ denotes the $h$-th head which is used to generate a certain granularity of phrase memory, and $F_{h}$ is a representation function with its own trainable parameters. The details for $F_{h}$ will be described in Section 3.2.

Then we perform attention on phrase level memory ${\bf H}^{g}$:
\begin{eqnarray}
    {\bf Q}^h, {\bf K}^h, {\bf V}^h &=& {\bf H}{\bf W}^h_{Q}, {\bf H}_{g}{\bf W}^h_{K}, {\bf H}_{g}{\bf W}^h_{V}\\
    {\bf O}^{h} &=& \textsc{Att}({\bf Q}^{h},{\bf K}^{h}) \ {\bf V}^{h}, 
\end{eqnarray}
where ${\bf Q}^h \in \mathbb{R}^{n \times d_{h}}, {\bf K}^h \in \mathbb{R}^{p \times d_{h}}, {\bf V}^h \in \mathbb{R}^{p \times d_{h}}$, the $p$ means the length of the key and value vectors which is decided by the granularity of phrase.  

Based on the single head self-attention, the final output of 
\textsc{Mg-Sa} can be expressed as follows: 
\begin{eqnarray}
    \textsc{Mg-Sa}(\bf H) = [{\bf O}^{1}, ..., {\bf O}^{N}], 
\end{eqnarray}
where $N$ denotes the number of heads. One head conducts either conventional word level attention or a certain granularity of phrase attention.

\subsection{Multi-Granularity Representation}

As seen in Figure~\ref{fig:mg-san}, multi-granularity phrases are simultaneously modeled by different heads.
To obtain the multi-granularity phrase representation, we first introduce phrase partition and composition strategies. Then, we describe phrase tag supervision and phrase interaction to further enhance the structure modeling on phrase representation.

\paragraph{Phrase Partition}
Partially inspired by \newcite{Shen:2018:ICLR}, we split the entire sequence into N-grams without overlaps. Such N-gram phrases are expressed as structurally adjacent and continuous items in the sequence. Formally, let ${\bf x}=(x_{1}, ..., x_{T})$ be a sequence, the phrases sequence of ${\bf x}$ can be denoted as is $P_{\bf x}=(p_{1}, ..., p_{M}), M=T/n$, where $p_{m}=(x_{n \times (m-1)}, ..., x_{n\times m}), 1 \leq m \leq M$, and $n$ denotes the length of the phrase which is a hyper-parameter. Padding is applied to the last phrase if necessary.

In addition, syntactic information has proven helpful in both SMT and NMT. We further introduce a syntactic phrase partition to represent well-formed structural information. 
Syntactic phrases organize words into nested constituents by using constituent parse tree. To obtain phrases in the view syntax, we break down the nodes at top K layers in the parse tree to capture top K levels of granularity for phrases, as illustrated in Figure~\ref{fig:mg-san} (a). Formally, one phrase in a certain layer of the parse tree can be defined as $p_{m}=(x^{1}, ..., x^{l})$, $l$ is the length of the phrase which is decided by the parse tree. The phrase sequence of the given input ${\bf x}$ is $P_{\bf x}=(p_{1}, ..., p_{M})$, $M$ is the number of phrase in the sequence.     

\paragraph{Composition Strategies}
Given phrase sequence $P_{\bf x}= (p_{1}, ..., p_{M})$ of input sequence, to capture local structure and context dependency inside each phrase and further generate phrase representation $Q_{M}$, we adopt phrase composition function to each phrase in the phrase sequence:
\begin{eqnarray}
     g_{m} = \textsc{Com}(p_{m}),
\end{eqnarray}
where \textsc{Com} is the composition function with shared parameters to all phrases, $g_{m} \in \mathbb{R}^{1\times d_{h}}$ is the phrase representation after composition.
There general choices of composition function are Convolution Neural Networks (\textsc{Cnn}s), Recurrent Neural Networks (\textsc{Rnn}s) and Self-attention Networks (\textsc{San}s). For \textsc{Cnn}s we only apply the Max-pooling layer. For \textsc{Rnn}s, we use the last hidden state of Long Short-term Memory Networks (\textsc{Lstm}) as phrase representation.  For \textsc{San}s, we use max pooling vector of the phrase to serve as the query for extracting inside phrase features to generate phrase representation. Then the phrase level memory of the input sequence can be denoted as ${\bf G}_{\bf x}=(g_{1}, ..., g_{M})$.

\paragraph{Phrase Tag Supervision}
Recent study shows auxiliary supervision on heads of \textsc{San}s can further improve semantic role labeling performance~\cite{Strubell:2018:EMNLP}. In this work, we leverage tag information as the auxiliary supervision on syntactic phrase representation. We argue that the proposed framework provide a natural way to incorporate syntactic tag signal of phrase representation. In detail, given phrase level memory ${\bf G}_{\bf x}=(g_{1}, ..., g_{M})$ after phrase composition, we predict the phrase tag of each phrase representation. We extract the node of each phrase in the constituent parsing tree to generate the phrase tag sequence ${\bf T}_{\bf x}=(t_{1}, ..., t_{M})$. $t_{i}$ denotes the tag for each phrase. For example, ``NP'' is the tag of the phrase ``a talk'' in second layer of parse tree, as shown in Figure~\ref{fig:mg-san} (a). We use the phrase representation to compute the probability of phrase tags:
\begin{eqnarray}
    p_{\theta_{i}} = softmax(W_{t}g_{i}+b_{t}), i= 1, ..., M,  
\end{eqnarray}
where $W_{t}$ and $b_{t}$ are parameters of tag generator. Formally, the phrase tag loss can be written as: 
\begin{eqnarray}
    \mathcal{L}_{tag} = -\sum_{i=1}^{M}t_{i}\log p_{\theta_{i}}(t_{i}). \label{eqn:tag}
\end{eqnarray}
The loss is equivalent to maximizing the conditional probability of tag sequence ${\bf T}_{\bf x}$ given phrase representation ${\bf G}_{\bf x}$.  

\paragraph{Phrase Interaction}
We introduce phrase interaction approach to better model dependencies between phrase representation. Since recurrence has proven important for capturing structure information~\cite{Tran:2018:EMNLP, Hao:2019:NAACL}, we propose to introduce recurrence to interact phrases and further model latent structure among phrases.
Specifically, we apply the recurrence function $\textsc{Rec}(\cdot)$ on the output of phrase composition ${\bf G}_{\bf x}=(g_{1}, ..., g_{M})$ in order to model the latent structure of the phrase sequence.
\begin{eqnarray}
     {\bf H}_{g} = \textsc{Rec}({\bf G}_{\bf x}),
\end{eqnarray}
where ${\bf H}_{g}$ is the final phrase level memory for the input layer $\bf H$. One general choice for $\textsc{Rec}(\cdot)$ is Long Short-term Memory Networks (\textsc{Lstm}).  
Recently, \newcite{Shen:2018:ICLR} introduce a new syntax-oriented inductive bias, namely ordered neurons, which enables LSTM models to perform tree-like composition without breaking its sequential form, and propose an advanced \textsc{Lstm} variant -- {\em Ordered Neurons} LSTM (\textsc{On-Lstm}).
\newcite{Hao:2019:EMNLPb} demonstrate the effectiveness of \textsc{On-Lstm} on modeling structure in NMT.
Accordingly, we further use \textsc{On-Lstm} for $\textsc{Rec}(\cdot)$, and expect \textsc{On-Lstm} can capture the latent structure under such syntax-oriented inductive bias between phrases. 

Finally, the representation function $F_{h}$ in Equation~\ref{eq:fh} of the framework can be summarized by the following components: 1). Phrase partition. 2). Phrase composition. 3). Phrase interaction.

\subsection{Training}
The training loss for a single training instance ${\bf x}=(x_{1}, ..., x_{T}), {\bf y}=(y_{1}, ..., y_{L})$ is defined as a weighted sum of the negative conditional log likelihood and the phrase tag loss. The total loss function can be written as:
\begin{equation}
    \mathcal{L} = -\sum_{i=1}^{L}y_{i}logP_{\omega}(y_{i}) + \lambda \mathcal{L}_{tag},
\end{equation}
where $\lambda$ is the coefficient to balance two loss functions and $\mathcal{L}_{tag}$ follows Equation~\ref{eqn:tag}. The hyperparameter $\lambda$ is empirically set to 0.001 in this work.

\section{Experiments}
In this section, we conduct experiments and make analysis to answer the following three questions:
\begin{itemize}
    \item[Q1.] Does the integration of the proposed \textsc{Mg-Sa} into the state-of-the-art \textsc{Transformer} improve the translation quality in terms of the BLEU score?
    \item[Q2.] Does the proposed \textsc{Mg-Sa} promote the generation of the target phrases?
    \item[Q3.] Does \textsc{Mg-Sa} capture more phrase information at the various granularity levels? 
\end{itemize}

In Section 4.1, we demonstrate that integrating the proposed \textsc{Mg-Sa} into the \textsc{Transformer} consistently improves the translation quality on both WMT14 English$\Rightarrow$German and NIST Chinese$\Rightarrow$English (Q1). Further analysis reveals that our approach has stronger ability of capturing the phrase information and promoting the generation of the target phrases (Q2).

In Section 4.2, we conduct experiments on the multi-granularity label prediction tasks~\cite{Shi:2016:EMNLP}, and investigate the representations of NMT encoders trained on both translation data and the training data of the label prediction tasks. Experimental results show that the proposed \textsc{Mg-Sa} indeed captures useful phrase information at various levels of granularities in both scenarios (Q3).

\subsection{Machine Translation}

\paragraph{Implementation Detail} 
We conduct the experiments on the WMT14 English-to-German (En$\Rightarrow$De) and NIST Chinese-to-English (Zh$\Rightarrow$En) translation tasks. 

For En$\Rightarrow$De,  the training dataset consists of 4.56M sentence pairs. We use the newstest2013 and newstest2014 as development set and test set respectively.
For Zh$\Rightarrow$En, the training dataset consists of about 1.25M sentence pairs. We used NIST MT02 dataset as development set, and  MT 03-06 datasets as test sets. Byte pair encoding (BPE) toolkit\footnote{https://github.com/rsennrich/subword-nmt}~\cite{sennrich2016neural} is used with 32K merge operations. We used case-sensitive NIST BLEU score~\cite{papineni:2002:ACL} as the evaluation metric, and bootstrap resampling~\cite{koehn2003statistical} for statistical significance test. We use the Stanford parser~\cite{klein2003accurate} to parse the sentences and obtain the relevant tags.

We test both {\em Base} and {\em Big} models, which differ at hidden size (512 vs. 1024), filter size (2048 vs. 4096) and the number of attention heads (8 vs. 16). All models are trained on eight NVIDIA Tesla P40 GPUs where each is allocated with a batch size of 4096 tokens. 
We implement the proposed approaches on top of \textsc{Transformer}~\cite{Vaswani:2017:NIPS} -- a state-of-the-art \textsc{San}s-based model on machine translation, and followed the setting in previous work~\cite{Vaswani:2017:NIPS} to train the models.

We incorporate the proposed model into the encoder. In each of our model variant, we maintain a quarter of heads for vanilla word level self-attention. For N-gram phrase models, we arrange the rest 3 quarters of heads for 2-gram, 3-gram and 4-gram respectively. For syntactic based models, we use the top 3 levels of granularity for phrases generated from constituent parse tree, each granularity of phrase modeled in a quarter of heads. There are many possible ways to implement the general idea of \textsc{Mg-Sa}. The aim of this paper is not to explore this whole space but simply to show that some fairly straightforward implementations work well.

Table~\ref{tab:phrase_partition},~\ref{tab:layers} and~\ref{tab:phrase} show the results on WMT14 English$\Rightarrow$German translation task with \textsc{Transformer-Base}. These results show the evaluation on the impact of different components.  

\paragraph{Phrase Composition} 
We investigate the effect of different phrase composition strategies with N-gram phrase partition. 
As seen in Table~\ref{tab:phrase_partition}, all proposed phrase composition methods consistently outperform \textsc{Transformer-Base} baseline, validating the importance of introducing multi-granularity phrase in \textsc{Transformer}.  Compared with other two models, \textsc{San}s achieve best performance with its strong representational powers inside the phrase, while only marginally increase the parameters and decrease the speed. We use \textsc{San}s phrase composition strategy as the default setting in subsequent experiments.  

\begin{table}[t]
\begin{center}
\begin{tabular}{c||c|c|c}
    {\bf Phrase Modeling}    &  {\bf \# Para.}   & {\bf Speed}&  {\bf BLEU}\\
    \hline
        n/a                &   88.0M   & 1.28 &  27.31\\ 
        \hline
           \textsc{Max-Pooling}       &  88.0M & 1.27 &   27.56\\
           \textsc{San}s        &  90.4M & 1.26 & \bf 27.69\\
           \textsc{Lstm}      &  96.1M & 1.14 &  27.58\\
\end{tabular}
\caption{Evaluation of various phrase composition strategies under N-gram phrase partition. ``\# Para'' denotes the trainable parameter size of each model (M=million), ``Speed'' denotes the training speed (steps/second).} 
  \label{tab:phrase_partition}
\end{center}
\end{table}

\begin{table}[t]
\begin{center}
\begin{tabular}{c||c|c|c}
  {\bf Encoder Layers}  &  {\bf \# Para.}   & {\bf Speed} &  {\bf BLEU} \\
    \hline
        $[1-6]$       &  90.4M  & 1.26&  27.69\\
        $[1-3]$        & 89.2M & 1.27 & 27.74\\
        $[1]$       & 88.4M  &  1.28 & \bf 27.83\\
\end{tabular}
\caption{Evaluation of different layers in the encoder, which are implemented as self-attention with \textsc{San}s phrase composition under N-gram partition. ``1'' denotes the bottom layer.}  
  \label{tab:layers}
\end{center}
\end{table}

\begin{table*}[t]
\begin{center}
\renewcommand\arraystretch{1.1}
\begin{tabular}{l|l||c|c||c|c}
      \# & {\bf Model Architecture}   & {\bf \# Para.}   &   {\bf Speed}  &   {\bf BLEU} & $\Delta$ \\
      \hline
      \hline
      1  &  \textsc{Transformer-Base} &  88.0M  & 1.28 & 27.31 & - \\
      \hline 
      2 & ~~~+ N-gram Phrase & 88.4M & 1.28 & 27.83 & +0.52 \\
      3 & ~~~+ Syntactic Phrase &  88.4M & 1.24 & 28.01 & +0.70 \\
      \cdashline{2-6}
      4 & ~~~+ Syntactic Phrase + $\mathcal{L}_{tag}$ & 88.4M & 1.23 & 28.07 & +0.76 \\
      \hline 
      5 & ~~~ ~~~ + \textsc{Lstm} Interaction & 89.5M  & 1.20 & 28.14 & +0.83 \\
      6 & ~~~ ~~~ + \textsc{On-Lstm} Interaction & 89.9M & 1.19 & {\bf 28.28} & +0.97 \\

\end{tabular}
\caption{Evaluation of phrase partition, tag supervision and interaction strategies.}
\label{tab:phrase}
\end{center}
\end{table*}

\paragraph{Encoder Layers}
Recent works ~\cite{Shi:2016:EMNLP, peters2018deep} show that different layers in encoder tend to capture different syntax and semantic features. Hence, there may have different needs for modeling phrase structure in each layer. In this experiment, we investigate the question of which layers should be applied with \textsc{Mg-Sa}. We apply \textsc{Mg-Sa} on different combination of layers. As shown in Table~\ref{tab:layers}, reducing the applied layers from high-level to low-level consistently increase translation quality in terms of BLEU score as well as the training speed. The results reveal that the bottom layer in encoder, which is directly taking word embedding as input, benefits more from modeling phrase structure. This phenomena verifies it is unnecessary to apply the phrase structure modeling to all layers. Accordingly, we only apply \textsc{Mg-Sa} in the bottom layer in the following experiments.

\paragraph{Phrase Partition and Tag Supervision}
As seen in Table~\ref{tab:phrase}, syntactic phrase partition (Row 3) improves the model performance over the N-gram phrase partition (Row 2), showing that the syntactic phrase benefits to translation quality. In addition, incorporating tag loss (Row 4) in training stage can further boost the translation performance. This indicates the auxiliary syntax objective is necessary, which is consistent with the results in other NLP task~\cite{Strubell:2018:EMNLP}. We use syntactic phrase partition with tag supervision as the default setting for subsequent experiments unless otherwise stated. 

\paragraph{Phrase Interaction}
As observed in Table~\ref{tab:phrase}, phrase interaction (Row 5-6) consistently improves performance of translation, proving the effectiveness and necessity of enhancing phrase level dependencies on phrase representation. \textsc{On-Lstm} based interaction (Row 6) outperforms its \textsc{Lstm} counterpart (Row 5). We attribute the improvement of \textsc{On-Lstm} to the stronger ability to perform syntax-oriented dependencies on phrase representation. We apply \textsc{On-Lstm} as the default setting for phrase interaction.

\paragraph{Main Results}
Table~\ref{tab:main} lists the results on WMT14 En$\Rightarrow$De and NIST Zh$\Rightarrow$En translation tasks. Our baseline models, outperform the reported results on the same data~\cite{Vaswani:2017:NIPS, zhang2019syntax}, which we believe make the evaluation convincing. As seen, in terms of BLEU score, the proposed \textsc{Mg-Sa} consistently improves translation performance across language pairs, which demonstrates the effectiveness and universality of the proposed approach. 

\begin{table*}[t]
\begin{center}
\begin{tabular}{l||rl|r l l l l l}
      \multirow{2}{*}{\bf Architecture}  & \multicolumn{2}{c|}{\bf En$\Rightarrow$De} & \multicolumn{6}{c}{\bf Zh$\Rightarrow$En}    \\
      \cline{2-9}
         &  \# Para.  &   BLEU &  \# Para. & MT03 &  MT04  & MT05 & MT06 & Avg  \\
    \hline \hline
    \multicolumn{9}{c}{{\em Existing NMT systems}} \\
    \hline
    \newcite{Vaswani:2017:NIPS} & 213M  & 28.4  & n/a  &  n/a & n/a  & n/a  & n/a  & n/a\\
    \newcite{zhang2019syntax}  & n/a & n/a & n/a &  40.45 & 42.76 & 40.09 & 39.67 & 40.74 \\ 
    \hline \hline 
    \multicolumn{9}{c}{{\em Our NMT systems}} \\
    \hline
        \textsc{Transformer-Base}  &        88.0M  &   27.31 & 73.4M &   41.88 & 44.48  & 42.21 & 41.93 & 42.60 \\
       ~~~  +\textsc{Mg-Sa}      & 89.9M  & 28.28$^\Uparrow$ & 75.3M &  43.98$^\Uparrow$  & 45.60$^\Uparrow$ & 44.28$^\Uparrow$ & 44.00$^\Uparrow$ & 44.46 \\ 
       \textsc{Transformer-Big}     & 264.1M & 28.58 & 234.8M  & 45.30 & 46.49 & 45.21 & 44.87 & 45.47 \\
         ~~~+\textsc{Mg-Sa}         & 271.5M & 29.01$^\uparrow$ & 242.2M & 45.76$^\uparrow$ & 46.81$^\uparrow$ & 45.77$^\uparrow$ & 46.48$^\Uparrow$ & 46.21
  \end{tabular}
  \caption{Comparing with the existing NMT systems on WMT14 En$\Rightarrow$De and NIST Zh$\Rightarrow$En test sets.  ``$\uparrow/\Uparrow$'': significant over the conventional self-attention counterpart ($p < 0.05/0.01$), tested by bootstrap resampling. ``\textsc{Mg-Sa}'' denotes ``Syntactic Phrase + $\mathcal{L}_{tag}$ + \textsc{On-Lstm} Interaction'' in Table~\ref{tab:phrase}.} 
  \label{tab:main}
  \end{center}
\end{table*}

\paragraph{Phrasal Pattern Evaluation}
\begin{figure}[t]
    \centering
    \includegraphics[width=0.45\textwidth]{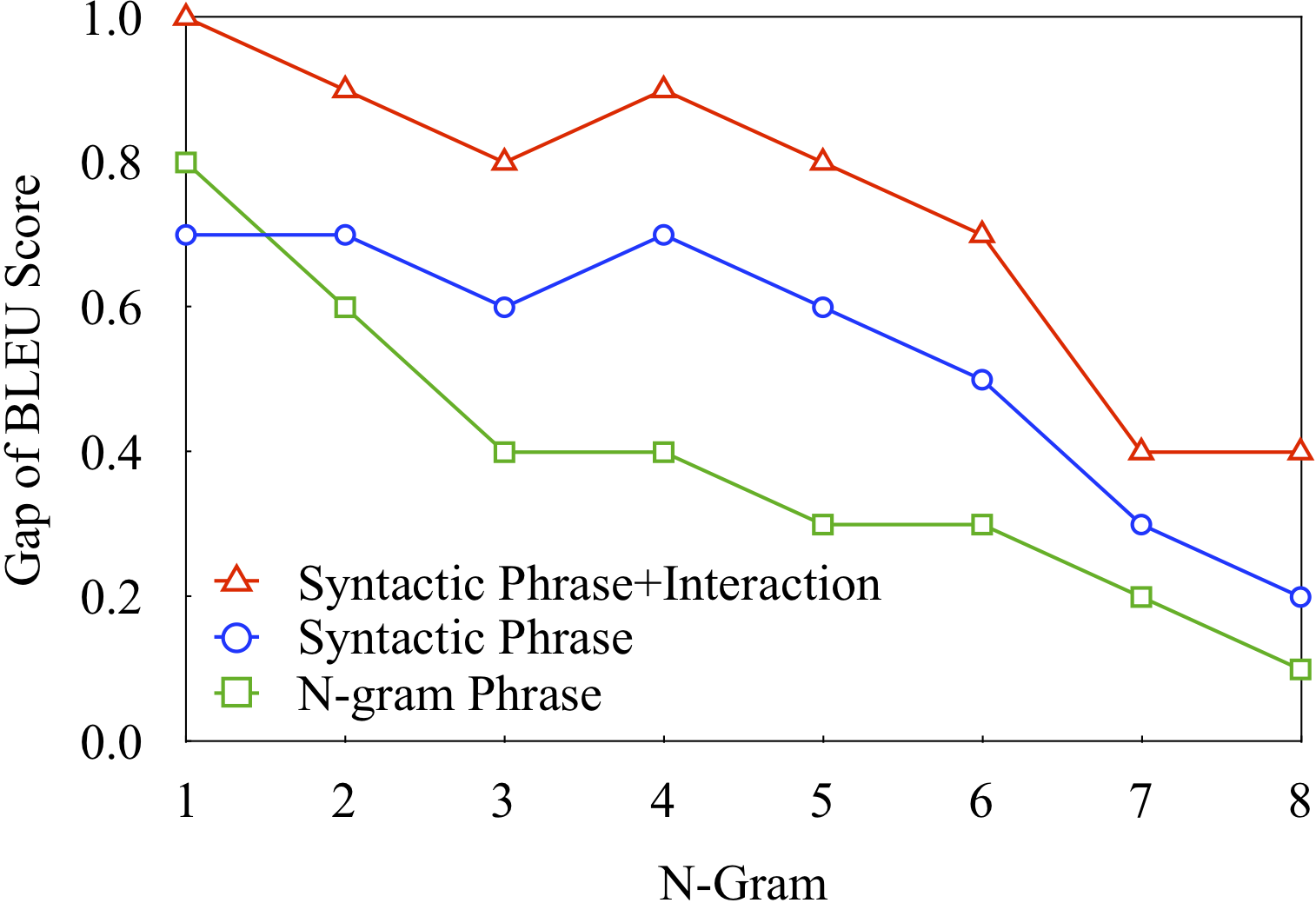}
    \caption{Performance improvement according to N-gram. Y-axis denotes the gap of BLEU score between our models and the baseline.}
    \label{fig:ngram_phrase}
\end{figure}

As aforementioned, the proposed \textsc{Mg-Sa} aims to simultaneously model different granularities of phrases with different heads in \textsc{San}s. 
To investigate whether the proposed \textsc{Mg-Sa} improves the generation of phrases in the output, we calculate the improvement of the proposed models over multiple N-grams, as shown in Figure~\ref{fig:ngram_phrase}. The results are reported on En$\Rightarrow$De validation set with \textsc{Transformer-Base}. 

Clearly, the proposed models consistently outperform the baseline model on all N-grams, indicating that the proposed \textsc{Mg-Sa} has stronger ability of capturing the phrase information and promoting the generation of the target phrases. Concerning the variations of proposed models, two syntactic phrase models outperforms the N-gram phrase model on larger phrases (i.e. 4-8 grams). We attribute this to the fact that more syntactic information is beneficial for the translation performance. This is also consistent with the strengths of phrase-based and syntax-based SMT models. 


\paragraph{Visualization of Attention}
\begin{figure*}[t]
    \centering
    \subfloat[Vanilla Multi-Head Self-Attention]{
    \includegraphics[height=0.35\textwidth]{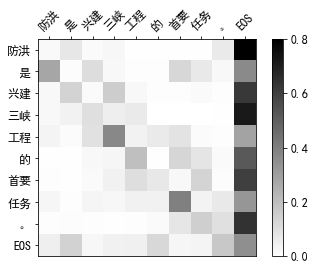}
    \label{fig:base}
    } \hspace{0.05\textwidth}
    \subfloat[Multi-Granularity Self-Attention]{
    \includegraphics[height=0.35\textwidth]{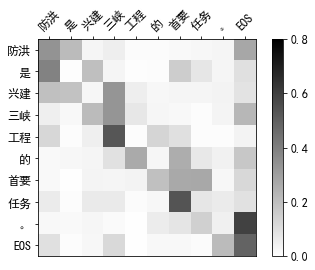}
    \label{fig:ours}
    }
    \caption{Visualization of attention examples of the same input sentence: (a) and (b) are produced by the vanilla multi-head self-attention and the proposed \textsc{Mg-Sa} models, respectively. Each row is the attention distribution over all the source tokens. The attention layer has 16 attention heads, and the attention weights in each row are the average of all the heads.}
    \label{fig:vis_att}
\end{figure*}

In order to evaluate whether the proposed model is able to capture phrase patterns or not, we visualize the attention layers in the encoder~\footnote{Since the attention weights of \textsc{Mg-Sa} cannot be visualized at word level, we visualize the attention weights in the subsequent layer after \textsc{Mh-Sa} and \textsc{Mg-Sa}.}. As shown in Fig.~\ref{fig:vis_att}, the vanilla model prefers to pay attention to the previous and next word and the end of the sentence, which is consistent with previous findings in ~\newcite{raganato2018analysis}. The proposed \textsc{Mg-Sa} successfully focuses on the phrases: 1) ``三峡 工程'', the 4th and the 5th rows in Fig.~\ref{fig:vis_att}(b), its English translation is `the Three Gorges Project'; 2) ``首要 任务'', the 7th and 8th rows in Fig.~\ref{fig:vis_att}(b), its English translation is `top priority'.  By visualizing the attention distributions, we believe the proposed \textsc{Mg-Sa} can capture phrase patterns to improve the translation performance.

\subsection{Multi-Granularity Phrases Evaluation}

\begin{table*}[t]
\begin{center}
\begin{tabular}{c|l|c c c c c|c}
        \# & {\bf Model} & \multicolumn{6}{c}{\bf Label Granularity: Large $\rightarrow$ Small}  \\
        \hline
       &   &  Voice & Tense  &  TSS & SPC & POS & Avg \\
     \hline
       \multicolumn{8}{c}{\em Pre-Trained NMT Encoder} \\
     \hline
     1& \textsc{Base} & 73.38  & 73.73 & 72.72 & 92.81 & 93.73 & 81.27 \\ 
     \hline
     2&  N-Gram Phrase & 73.06 &72.83 & 72.11 &96.42& 96.34 & 82.15 \\
     3& Syntactic Phrase &73.37 &73.62& 75.60 & 96.72 & 96.68 & 83.19\\
     4&  Syntactic Phrase + Interaction  & 73.20 & 74.78 & 75.24 &96.78 & 96.56 & 83.31\\
    \hline
      \multicolumn{8}{c}{\em Train From Scratch} \\
    \hline
        5&    \textsc{Base} & 83.46 & 85.39 & 83.44  & 96.35 & 96.12 & 88.95\\
    \hline
        6& N-Gram Phrase & 83.55 & 85.62 & 85.21  & 96.23 & 96.17 & 89.36\\
       7& Syntactic Phrase & 84.70 &  87.52 &  97.42 & 96.95  & 96.24 & 92.57 \\
       8& Syntactic Phrase + Interaction & 86.45 & 87.65 &  99.07 & 96.99 & 96.40 & 93.31 \\ 
    
\end{tabular}
\caption{Accuracies on multi-granularity label prediction tasks. ``Pre-Trained NMT Encoder'' denotes using the pre-trained NMT encoders of model variations in Table~\ref{tab:phrase}. ``Train From Scratch'' denotes using three encoder layers with proposed \textsc{Mg-Sa} variants, which are trained from scratch. For syntactic phrase based models, we only apply syntactic boundary of phrases and do not use any tag supervision for fair comparison. } 
\label{tab:tasks}
\end{center}
\end{table*}

In this section, we conduct multi-granularity label prediction tasks to the proposed models in terms of whether the proposed model is effective as expected to capture different levels of granularity phrase information of sentences. We analyze the impact of multi-granularity self-attention based on two sets of experiments. The first set of experiments are probing the pre-trained NMT encoders, which aims to evaluate the linguistics knowledge embedded in the NMT encoder output in the machine translation section. Furthermore, to test the ability of the \textsc{Mg-Sa} itself, we conduct the second set of experiments, which are on the same tasks using encoder models trained from scratch.  

\paragraph{Tasks}
\newcite{Shi:2016:EMNLP} propose 5 tasks to predict various granularity syntactic labels of from sentence to word in order to investigate whether an encoder can learn syntax information. These labels are: ``Voice'': active or passive, ``Tense'': past or non-past of main-clause verb, ``TSS'': top-level syntactic sequence of constituent tree, and two word-level syntactic label tasks, ``SPC'': the smallest phrase constituent that above each word, ``POS'': Part-of-Speech tags for each words. The tasks for predicting larger labels require models to capture and record larger granularity of phrase information of sentences~\cite{Shi:2016:EMNLP}. We conduct these tasks to study whether the proposed \textsc{Mg-Sa} benefits the multi-granularity phrase modeling to produce more useful and informative representation. 

\paragraph{Data and Models}
We extracted the sentences from the Toronto Book Corpus~\cite{zhu2015aligning}.
We sample and pre-process 120k sentences for each task following~\newcite{conneau:2018:acl}. By instruction of \newcite{Shi:2016:EMNLP}, we label these sentences for each task. The train/valid/test dataset ratios are set to 10/1/1. 

For pre-trained NMT encoders, we use the pre-trained encoders of model variations in Table~\ref{tab:phrase} followed by a MLP classifier, which are used to carry out probing tasks. 

For models trained from scratch, each of our model consists of 3 encoding layers followed by a MLP classifier. For each encoding layer, we employ a multi-head self-attention block and a feed-forward block as in \textsc{Transformer}, which have shown significant performance on several NLP tasks~\cite{Devlin:2019:NAACL}. The difference between the compared models merely lies in the self-attention mechanism: ``\textsc{Base}'' denotes standard \textsc{Mh-Sa}, ``N-Gram Phrase'' and ``Syntactic Phrase'' are the proposed \textsc{Mg-Sa} under N-gram phrase and syntactic phrase partition, and ``Syntactic Phrase + Interaction'' denotes \textsc{Mg-Sa} with phrase interaction by using \textsc{On-Lstm}. We use  same assignments of heads for multi-granularity phrases as machine translation task for all model variants. 

\paragraph{Results Analysis}
Table~\ref{tab:tasks} lists the prediction accuracies of five syntactic labels on test. Several observations can be made here.
1). Comparing the two set of experiments, the experimental results from models trained from scratch consistently outperform the results from NMT encoder probing on all tasks.
2). The models with syntactic information (Rows 3-4, 7-8) significantly perform better than those models without incorporating syntactic information (Rows 1-2, 5-6).
3). For NMT probing, the proposed models outperform the baseline model especially on relative small granularity of phrases information, such as `SPC' and `POS' tasks.  
4). If trained from scratch, the proposed models achieve more improvements on predicting larger granularities of labels, such as `TSS', `Tense' and `Voice' tasks, which require models to record larger phrase of sentences~\cite{Shi:2016:EMNLP}. The results show that the applicability of the proposed \textsc{Mg-Sa} is not limited to machine translation, but also on monolingual tasks. 

\section{Related Works}
\paragraph{Phrase Modeling for NMT}
Several works have proven that the introduction of phrase modeling in NMT can obtain promising improvement on translation quality. Tree-based encoders, which explicitly take the constituent tree~\cite{eriguchi:2016:acl} or dependency tree~\cite{bastings-etal-2017-graph}  into consideration, are proposed to produce tree-based phrase representations. The difference of our work from these studies is that they adopt the RNN-based encoder to form the tree-based encoder while we explicitly introduce the phrase structure into the the state-of-the-art multi-layer multi-head \textsc{San}s-based encoder, which we believe is more challenging.

Another thread of work is to implicitly promote the generation of phrase-aware representation, such as the integration of external phrase boundary~\cite{wang:2017:emnlp, nguyen:2018:arxiv, li2018area}, prior attention bias~\cite{yang:2018:emnlp,yang2019convolutional,guo2019gaussian}. Our work differs at that we explicitly model phrase patterns at different granularities, which is then attended by different attention heads.

\paragraph{Multi Granularity Representation}
Multi-granularity representation, which is proposed to make full use of subunit composition at different levels of granularity, has been explored in various NLP tasks, such as paraphrase identification~\cite{yin2015convolutional}, Chinese word embedding learning~\cite{yin2016multi}, universal sentence encoding~\cite{wu:2018:emnlp} and machine translation~\cite{nguyen:2018:arxiv, li2018area}. The major difference between our work and \newcite{nguyen:2018:arxiv,li2018area} lies in that we successfully introduce syntactic information into our multi-granularity representation. Furthermore, it is not well measured how much phrase information are stored in multi-granularity representation. We conduct the multi-granularity label prediction tasks and empirically verify that the phrase information is embedded in the multi-granularity representation.

\paragraph{Multi-Head Attention}
Multi-head attention mechanism has shown its effectiveness in machine translation~\cite{Vaswani:2017:NIPS} and generative dialog~\cite{tao2018get} systems. 
Recent studies shows that the modeling ability of multi-head attention has not been completely developed. Several specific guidance cues of different heads without breaking the vanilla multi-head attention mechanism can further boost the performance, e.g., disagreement regularization~\cite{Li:2018:EMNLP,tao2018get}, information aggregation~\cite{li2019information}, and functional specialization~\cite{Fan:2019:ACL} on attention heads,  the combination of multi-head attention with multi-task learning~\cite{Strubell:2018:EMNLP}. Our work demonstrates that multi-head attention also benefits from the integration of the phrase information.

\section{Conclusion}
In this paper, we propose multi-granularity self-attention model, a novel attention mechanism to simultaneously attend different granularity phrase. We study effective phrase representation for N-gram phrase and syntactic phrase, and find that a syntactic phrase based mechanism obtains the best result due to effectively incorporating rich syntactic information. To evaluate the effectiveness of the proposed model, we conduct experiments on widely-used WMT14 En$\Rightarrow$De and NIST Zh$\Rightarrow$En datasets. Experimental results on two language pairs show that the proposed model achieve significant improvements over the baseline \textsc{Transformer}. Targeted multi-granularity phrases evaluation shows that our model indeed capture useful phrase information.  

As our approach is not limited to specific tasks, it is interesting to validate the proposed model in other tasks, such as reading comprehension, language inference, and sentence classification. 

\section*{Acknowledgments}
J.Z. was supported by the National Institute of General Medical Sciences of the National Institute of Health under award number R01GM126558. 
We thank the anonymous reviewers for their insightful comments.

\bibliography{emnlp-ijcnlp-2019}

\begin{thebibliography}{36}
\expandafter\ifx\csname natexlab\endcsname\relax\def\natexlab#1{#1}\fi

\bibitem[{Bastings et~al.(2017)Bastings, Titov, Aziz, Marcheggiani, and
  Simaan}]{bastings-etal-2017-graph}
Joost Bastings, Ivan Titov, Wilker Aziz, Diego Marcheggiani, and Khalil Simaan.
  2017.
\newblock Graph convolutional encoders for syntax-aware neural machine
  translation.
\newblock In \emph{EMNLP}.

\bibitem[{Chiang(2005)}]{chiang:2005:acl}
David Chiang. 2005.
\newblock A hierarchical phrase-based model for statistical machine
  translation.
\newblock In \emph{ACL}.

\bibitem[{Conneau et~al.(2018)Conneau, Kruszewski, Lample, Barrault, and
  Baroni}]{conneau:2018:acl}
Alexis Conneau, German Kruszewski, Guillaume Lample, Lo{\"\i}c Barrault, and
  Marco Baroni. 2018.
\newblock What you can cram into a single ${\$}{\&}!{\#}*$ vector: Probing
  sentence embeddings for linguistic properties.
\newblock In \emph{ACL}.

\bibitem[{Devlin et~al.(2019)Devlin, Chang, Lee, and
  Toutanova}]{Devlin:2019:NAACL}
Jacob Devlin, Ming-Wei Chang, Kenton Lee, and Kristina Toutanova. 2019.
\newblock Bert: Pre-training of deep bidirectional transformers for language
  understanding.
\newblock In \emph{NAACL}.

\bibitem[{Eriguchi et~al.(2016)Eriguchi, Hashimoto, and
  Tsuruoka}]{eriguchi:2016:acl}
Akiko Eriguchi, Kazuma Hashimoto, and Yoshimasa Tsuruoka. 2016.
\newblock Tree-to-sequence attentional neural machine translation.
\newblock In \emph{ACL}.

\bibitem[{Fan et~al.(2019)Fan, Lewis, and Dauphin}]{Fan:2019:ACL}
Angela Fan, Mike Lewis, and Yann Dauphin. 2019.
\newblock Strategies for structuring story generation.
\newblock In \emph{ACL}.

\bibitem[{Guo et~al.(2019)Guo, Zhang, and Liu}]{guo2019gaussian}
Maosheng Guo, Yu~Zhang, and Ting Liu. 2019.
\newblock Gaussian transformer: a lightweight approach for natural language
  inference.
\newblock In \emph{AAAI}.

\bibitem[{Hao et~al.(2019{\natexlab{a}})Hao, Wang, Shi, Zhang, and
  Tu}]{Hao:2019:EMNLPb}
Jie Hao, Xing Wang, Shuming Shi, Jinfeng Zhang, and Zhaopeng Tu.
  2019{\natexlab{a}}.
\newblock Towards better modeling hierarchical structure for self-attention
  with ordered neurons.
\newblock In \emph{EMNLP}.

\bibitem[{Hao et~al.(2019{\natexlab{b}})Hao, Wang, Yang, Wang, Zhang, and
  Tu}]{Hao:2019:NAACL}
Jie Hao, Xing Wang, Baosong Yang, Longyue Wang, Jinfeng Zhang, and Zhaopeng Tu.
  2019{\natexlab{b}}.
\newblock Modeling recurrence for transformer.
\newblock In \emph{NAACL}.

\bibitem[{Klein and Manning(2003)}]{klein2003accurate}
Dan Klein and Christopher~D Manning. 2003.
\newblock Accurate unlexicalized parsing.
\newblock In \emph{ACL}.

\bibitem[{Koehn et~al.(2003)Koehn, Och, and Marcu}]{koehn2003statistical}
Philipp Koehn, Franz~Josef Och, and Daniel Marcu. 2003.
\newblock Statistical phrase-based translation.
\newblock In \emph{NAACL}.

\bibitem[{Li et~al.(2018)Li, Tu, Yang, Lyu, and Zhang}]{Li:2018:EMNLP}
Jian Li, Zhaopeng Tu, Baosong Yang, Michael~R. Lyu, and Tong Zhang. 2018.
\newblock {Multi-Head Attention with Disagreement Regularization}.
\newblock In \emph{EMNLP}.

\bibitem[{Li et~al.(2019{\natexlab{a}})Li, Yang, Dou, Wang, Lyu, and
  Tu}]{li2019information}
Jian Li, Baosong Yang, Zi-Yi Dou, Xing Wang, Michael~R Lyu, and Zhaopeng Tu.
  2019{\natexlab{a}}.
\newblock Information aggregation for multi-head attention with
  routing-by-agreement.
\newblock In \emph{NAACL}.

\bibitem[{Li et~al.(2019{\natexlab{b}})Li, Kaiser, Bengio, and Si}]{li2018area}
Yang Li, Lukasz Kaiser, Samy Bengio, and Si~Si. 2019{\natexlab{b}}.
\newblock Area attention.
\newblock In \emph{ICML}.

\bibitem[{Liu et~al.(2006)Liu, Liu, and Lin}]{Liu:2006:ACL}
Yang Liu, Qun Liu, and Shouxun Lin. 2006.
\newblock Tree-to-string alignment template for statistical machine
  translation.
\newblock In \emph{ACL}.

\bibitem[{Nguyen and Joty(2018)}]{nguyen:2018:arxiv}
Phi~Xuan Nguyen and Shafiq Joty. 2018.
\newblock Phrase-based attentions.
\newblock \emph{arXiv preprint arXiv:1810.03444}.

\bibitem[{Papineni et~al.(2002)Papineni, Roukos, Ward, and
  Zhu}]{papineni:2002:ACL}
Kishore Papineni, Salim Roukos, Todd Ward, and Wei-Jing Zhu. 2002.
\newblock Bleu: a method for automatic evaluation of machine translation.
\newblock In \emph{ACL}.

\bibitem[{Peters et~al.(2018)Peters, Neumann, Iyyer, Gardner, Clark, Lee, and
  Zettlemoyer}]{peters2018deep}
Matthew~E Peters, Mark Neumann, Mohit Iyyer, Matt Gardner, Christopher Clark,
  Kenton Lee, and Luke Zettlemoyer. 2018.
\newblock Deep contextualized word representations.
\newblock In \emph{NAACL}.

\bibitem[{Raganato and Tiedemann(2018)}]{raganato2018analysis}
Alessandro Raganato and J{\"o}rg Tiedemann. 2018.
\newblock An analysis of encoder representations in transformer-based machine
  translation.
\newblock In \emph{EMNLP Workshop BlackboxNLP: Analyzing and Interpreting
  Neural Networks for NLP}.

\bibitem[{Sennrich et~al.(2016)Sennrich, Haddow, and
  Birch}]{sennrich2016neural}
Rico Sennrich, Barry Haddow, and Alexandra Birch. 2016.
\newblock Neural machine translation of rare words with subword units.
\newblock In \emph{ACL}.

\bibitem[{Shen et~al.(2018)Shen, Zhou, Long, Jiang, and Zhang}]{Shen:2018:ICLR}
Tao Shen, Tianyi Zhou, Guodong Long, Jing Jiang, and Chengqi Zhang. 2018.
\newblock Bi-directional block self-attention for fast and memory-efficient
  sequence modeling.
\newblock In \emph{ICLR}.

\bibitem[{Shi et~al.(2016)Shi, Padhi, and Knight}]{Shi:2016:EMNLP}
Xing Shi, Inkit Padhi, and Kevin Knight. 2016.
\newblock Does string-based neural mt learn source syntax.
\newblock In \emph{EMNLP}.

\bibitem[{Strubell et~al.(2018)Strubell, Verga, Andor, Weiss, and
  McCallum}]{Strubell:2018:EMNLP}
Emma Strubell, Patrick Verga, Daniel Andor, David Weiss, and Andrew McCallum.
  2018.
\newblock {Linguistically-Informed Self-Attention for Semantic Role Labeling}.
\newblock In \emph{EMNLP}.

\bibitem[{Tao et~al.(2018)Tao, Gao, Shang, Wu, Zhao, and Yan}]{tao2018get}
Chongyang Tao, Shen Gao, Mingyue Shang, Wei Wu, Dongyan Zhao, and Rui Yan.
  2018.
\newblock Get the point of my utterance! learning towards effective responses
  with multi-head attention mechanism.
\newblock In \emph{IJCAI}.

\bibitem[{Tran et~al.(2018)Tran, Bisazza, and Monz}]{Tran:2018:EMNLP}
Ke~Tran, Arianna Bisazza, and Christof Monz. 2018.
\newblock The importance of being recurrent for modeling hierarchical
  structure.
\newblock In \emph{EMNLP}.

\bibitem[{Vaswani et~al.(2017)Vaswani, Shazeer, Parmar, Uszkoreit, Jones,
  Gomez, Kaiser, and Polosukhin}]{Vaswani:2017:NIPS}
Ashish Vaswani, Noam Shazeer, Niki Parmar, Jakob Uszkoreit, Llion Jones,
  Aidan~N Gomez, {\L}ukasz Kaiser, and Illia Polosukhin. 2017.
\newblock {Attention is All You Need}.
\newblock In \emph{NIPS}.

\bibitem[{Voita et~al.(2019)Voita, Talbot, Moiseev, Sennrich, and
  Titov}]{Voita:2019:ACL}
Elena Voita, David Talbot, Fedor Moiseev, Rico Sennrich, and Ivan Titov. 2019.
\newblock Analyzing multi-head self-attention: Specialized heads do the heavy
  lifting, the rest can be pruned.
\newblock In \emph{ACL}.

\bibitem[{Wang et~al.(2017)Wang, Tu, Xiong, and Zhang}]{wang:2017:emnlp}
Xing Wang, Zhaopeng Tu, Deyi Xiong, and Min Zhang. 2017.
\newblock Translating phrases in neural machine translation.
\newblock In \emph{EMNLP}.

\bibitem[{Wu et~al.(2018)Wu, Wang, Liu, and Ma}]{wu:2018:emnlp}
Wei Wu, Houfeng Wang, Tianyu Liu, and Shuming Ma. 2018.
\newblock Phrase-level self-attention networks for universal sentence encoding.
\newblock In \emph{EMNLP}.

\bibitem[{Yang et~al.(2018)Yang, Tu, Wong, Meng, Chao, and
  Zhang}]{yang:2018:emnlp}
Baosong Yang, Zhaopeng Tu, Derek~F. Wong, Fandong Meng, Lidia~S. Chao, and Tong
  Zhang. 2018.
\newblock Modeling localness for self-attention networks.
\newblock In \emph{EMNLP}.

\bibitem[{Yang et~al.(2019)Yang, Wang, Wong, Chao, and
  Tu}]{yang2019convolutional}
Baosong Yang, Longyue Wang, Derek Wong, Lidia~S Chao, and Zhaopeng Tu. 2019.
\newblock Convolutional self-attention networks.
\newblock In \emph{NAACL}.

\bibitem[{Yin et~al.(2016)Yin, Wang, Li, Li, and Wang}]{yin2016multi}
Rongchao Yin, Quan Wang, Peng Li, Rui Li, and Bin Wang. 2016.
\newblock Multi-granularity chinese word embedding.
\newblock In \emph{EMNLP}.

\bibitem[{Yin and Sch{\"u}tze(2015)}]{yin2015convolutional}
Wenpeng Yin and Hinrich Sch{\"u}tze. 2015.
\newblock Convolutional neural network for paraphrase identification.
\newblock In \emph{NAACL}.

\bibitem[{Zhang et~al.(2019)Zhang, Li, Fu, and Zhang}]{zhang2019syntax}
Meishan Zhang, Zhenghua Li, Guohong Fu, and Min Zhang. 2019.
\newblock Syntax-enhanced neural machine translation with syntax-aware word
  representations.
\newblock In \emph{NAACL}.

\bibitem[{Zhao et~al.(2018)Zhao, Wang, Zhang, and Zong}]{czhaophrase}
Yang Zhao, Yining Wang, Jiajun Zhang, and Chengqing Zong. 2018.
\newblock Phrase table as recommendation memory for neural machine translation.
\newblock In \emph{IJCAI}.

\bibitem[{Zhu et~al.(2015)Zhu, Kiros, Zemel, Salakhutdinov, Urtasun, Torralba,
  and Fidler}]{zhu2015aligning}
Yukun Zhu, Ryan Kiros, Rich Zemel, Ruslan Salakhutdinov, Raquel Urtasun,
  Antonio Torralba, and Sanja Fidler. 2015.
\newblock Aligning books and movies: Towards story-like visual explanations by
  watching movies and reading books.
\newblock In \emph{ICCV}.

\end{thebibliography}
\bibliographystyle{acl_natbib}
\end{CJK}
\end{document}